\documentclass[11pt,a4paper]{article}
\usepackage[hyperref]{emnlp-ijcnlp-2019}
\usepackage{times}
\usepackage{latexsym}
\usepackage{multirow}
\usepackage{graphicx}
\usepackage{pgfplots}
\usepackage{amsmath,amssymb}
\usepackage{algorithm}
\usepackage[noend]{algpseudocode}
\algrenewcommand\algorithmicrequire{\textbf{Input:}}
\algrenewcommand\algorithmicensure{\textbf{Output:}}
\pgfplotsset{compat=1.14}

\usepackage{url}

\aclfinalcopy % Uncomment this line for the final submission

\newcommand\geo{\textsc{GeoQuery}}
\newcommand\scholar{\textsc{Scholar}}
\newcommand\overnight{\textsc{Overnight}}
\newcommand\detect{\textsc{GrAnno}}
\newcommand\md{$D$}
\newcommand\dnat{\sD_{\text{nat}}}
\newcommand\don{\sD_{\text{on}}}
\newcommand\dlang{\sD_{\text{lang}}}
\newcommand\dga{\sD_{\text{ga}}}
\newcommand\cnat{\sC_{\text{nat}}}
\newcommand\con{\sC_{\text{on}}}
\newcommand\xul{\sX_{\text{ul}}}
\newcommand\comment[1]{}
\newcommand{\emailsize}{\fontsize{12pt}\times}

\newcommand\nl[1]{{\it``#1''}} % Natural language
\newcommand\zl[1]{\text{\footnotesize{\tt #1}}} % Logical Language (in the world)
\newcommand\sC{\ensuremath{\mathcal{C}}}
\newcommand\sD{\ensuremath{\mathcal{D}}}

\newcommand\sS{\ensuremath{\mathcal{S}}}

\newcommand\sX{\ensuremath{\mathcal{X}}}

\title{\emph{Don't paraphrase, detect!} Rapid and Effective Data Collection for Semantic Parsing}

\author{Jonathan Herzig$^{1}$  ~~~~~Jonathan Berant$^{1,2}$ \\
\mbox{}\\
$^1$School of Computer Science, Tel-Aviv University \\
$^2$Allen Institute for Artificial Intelligence \\
\emailsize{\texttt{\{jonathan.herzig,joberant\}@cs.tau.ac.il}}}
\date{}

\begin{document}
\maketitle
\begin{abstract}
A major hurdle on the road to conversational interfaces is the difficulty in collecting data that maps language utterances to logical forms. One prominent approach for data collection has been to automatically generate pseudo-language paired with logical forms, and paraphrase the pseudo-language to natural language through crowdsourcing \cite{wang2015overnight}. However, this data collection procedure often leads to low performance on real data, due to a mismatch between the true distribution of examples and the distribution induced by the data collection procedure. In this paper, we thoroughly analyze two sources of mismatch in this process:
the mismatch in \emph{logical form distribution} and the mismatch in \emph{language distribution} between the true and induced distributions.
We quantify the effects of these mismatches, and propose a new data collection approach that mitigates them. Assuming access to unlabeled utterances from the true distribution, we combine crowdsourcing with a paraphrase model to detect correct logical forms for the unlabeled utterances. On two datasets, our method leads to 70.6 accuracy on average on the \emph{true distribution}, compared to 51.3 in  paraphrasing-based data collection. 
\end{abstract}

\section{Introduction}
\label{introduction}

\begin{figure}[t]
  \includegraphics[width=1.0\columnwidth]{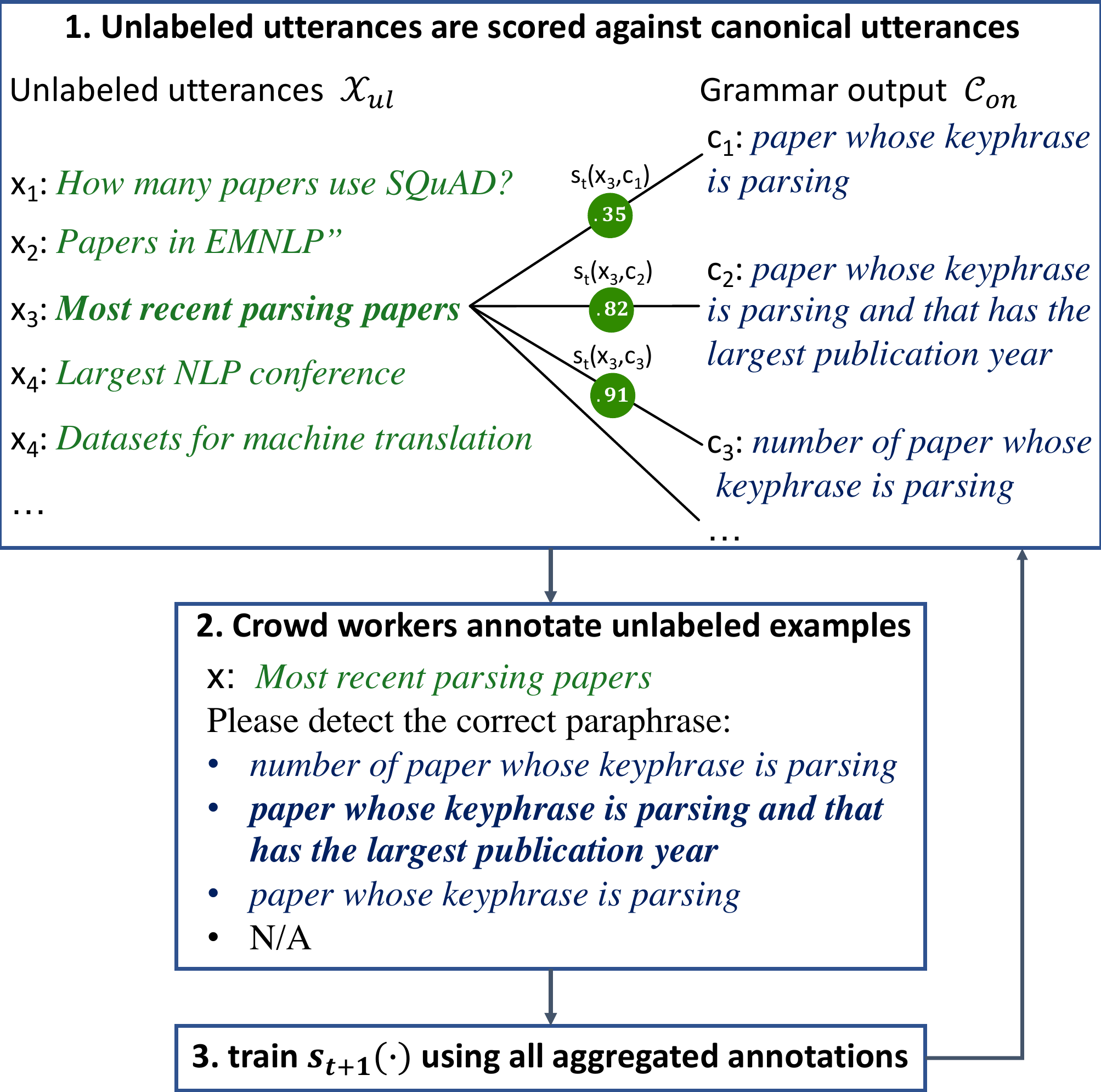}
  \caption{An overview of \detect{}, a method for annotating unlabeled utterances with their logical forms.
  }~\label{fig:detect}
\end{figure}

Conversing with a virtual assistant in natural language is one of the most exciting current applications of semantic parsing, the task of mapping natural language utterances to executable logical forms \cite{zelle96geoquery, zettlemoyer05ccg, liang11dcs}. 
Semantic parsing models rely on supervised training data that pairs natural language utterances with logical forms. Alas, such data does not occur naturally, especially in virtual assistants that are meant to support thousands of different applications and use-cases. Thus, efficient data collection is perhaps the most pressing problem for scalable conversational interfaces.

In recent years, many attempts aim to reduce the burden of data collection for semantic parsing, including training from denotations \cite{kwiatkowski2013scaling,artzi2013weakly}, semi-supervised learning \cite{kocisk2016semantic, yin-etal-2018-structvae}, human in the loop \cite{iyer2017neural,lawrence-riezler-2018-improving}, and training on examples from other domains \cite{herzig2017multi,herzig2018zeroshot,su2017cross}.
One prominent approach for data collection was introduced by \newcite{wang2015overnight}, termed \overnight{}. In this method, one automatically generates logical forms for an application from a grammar, paired with pseudo-language utterances. These pseudo-language utterances are then shown to crowd workers, who are able to understand them and paraphrase them into natural language, resulting in a supervised dataset.

The \overnight{} procedure has been adopted and extended both within semantic parsing
%\overnight{} has been adopted and extended both within semantic parsing
\cite{locascio-etal-2016-neural,ravichander-etal-2017-say,zhongSeq2SQL2017,cheng2018building}, in dialogue \cite{shah2018building, damonte2019practical} and in visual reasoning \cite{johnson2017clevr,hudson2019gqa}.

While the \overnight{} approach is appealing since it generates training data from scratch, it suffers from a major drawback -- training a parser on data generated from \overnight{} and testing on utterances collected from a target distribution results in a significant drop in performance \cite{wang2015overnight,ravichander-etal-2017-say}.

In this paper, we thoroughly analyze the sources of mismatch between a target distribution and the distribution induced by \overnight{}, and propose a new method for overcoming this mismatch. We decouple the mismatch into two terms: the \emph{logical form mismatch}, i.e., the difference between the target distribution of logical forms and the distribution obtained when generating from a grammar, and the \emph{language mismatch}, i.e., the difference between the natural language obtained when paraphrasing a pseudo-language utterance and the language obtained by real users of an application.

We find that the most severe problem arising from the \emph{logical form mismatch}
is insufficient coverage of logical form templates that occur in the true distribution, when generating from a grammar. We also isolate the \emph{language mismatch} effect by paraphrasing logical forms sampled from the true logical form distribution, and find that the language mismatch alone results in a decrease of 9 accuracy points on average on two datasets.

To overcome these mismatches, we propose an alternative method to \overnight{}, that utilizes unlabeled utterances. Our method, named \detect{} (grammar-driven annotation), allows crowd workers to iteratively annotate unlabeled utterances by \emph{detecting} their correct grammar generated paraphrase. Figure \ref{fig:detect} illustrates a single iteration of this approach. An unlabeled utterance is matched using a paraphrase model against candidate pseudo-language utterances generated by the grammar (step 1). The unlabeled utterance and its top candidate paraphrases are presented to a crowd worker that detects the correct paraphrase (step 2). The paraphrase model is re-trained given all annotated utterances thus far (step 3), and is used to score the remaining unlabeled utterances.

On two semantic parsing datasets, we show our procedure leads to annotation of 89\% of the original training data. The accuracy of the resulting parser is 70.6 on average, well beyond the accuracy obtained through the original \overnight{} procedure at 51.3. This substantially closes the gap to a fully-supervised semantic parser, which obtains 84.9 accuracy.
All our code and collected data is available at \url{https://github.com/jonathanherzig/semantic-parsing-annotation}.
\section{The \overnight{} Framework}
\label{background}

\begin{figure}[t]
  \includegraphics[width=1.0\columnwidth]{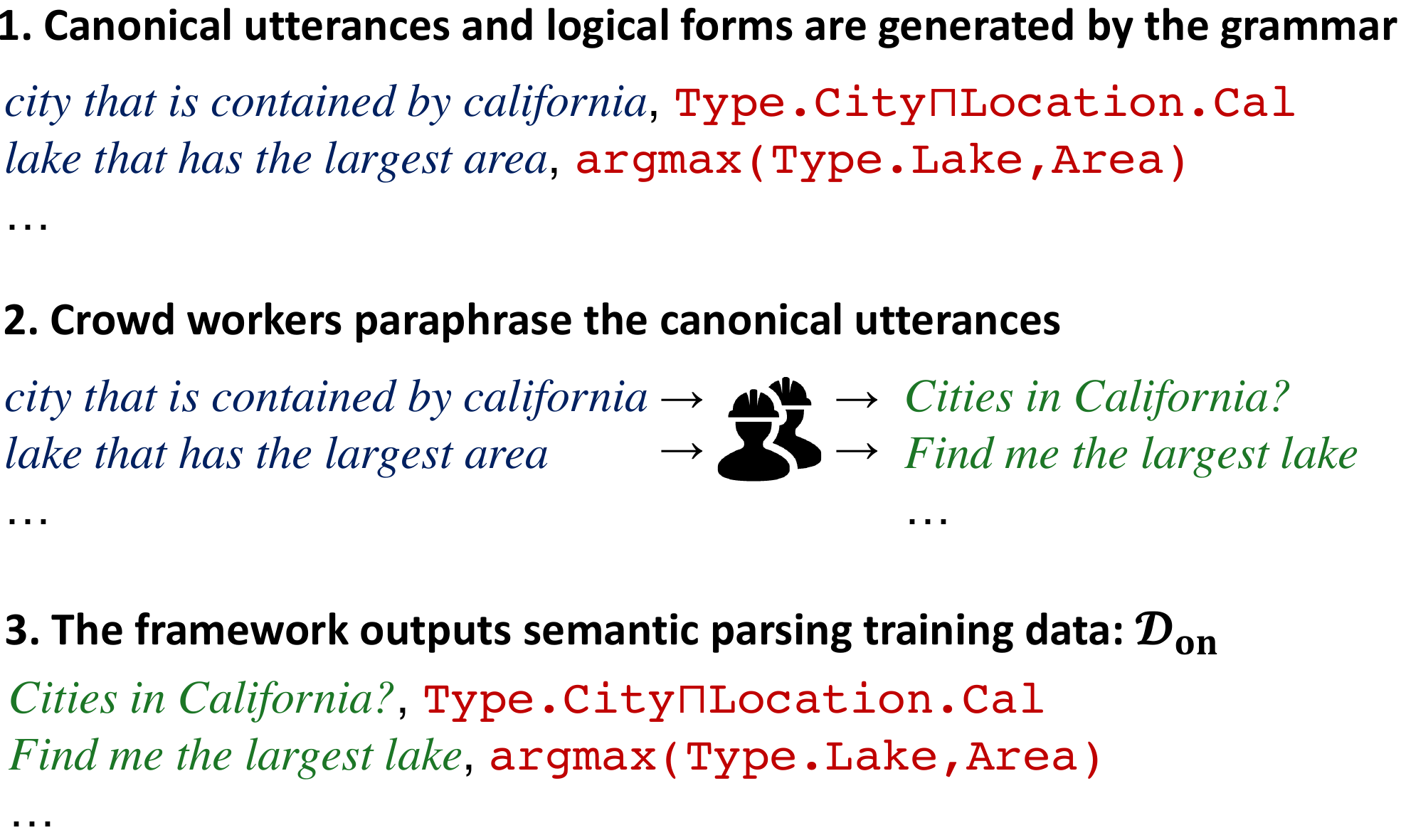}
  \caption{ \overnight{}: Canonical utterances are generated by a grammar and paraphrased by crowd workers.
  }~\label{fig:overnight}
\end{figure}

We now describe the \overnight{} framework for data collection, which we
investigate and improve upon in this work. The starting point is a user who needs a
semantic parser for some domain, but has \emph{no
data}. \overnight{} describes a procedure for generating training data from
scratch, which comprises two steps (see Figure~\ref{fig:overnight}). 
First, a synchronous grammar is used to generate logical forms paired with  
\emph{canonical utterances}, which are pseudo-language utterances that are
understandable to people, but do not sound natural. Second, crowd workers
paraphrase these canonical utterances into natural language utterances. This
results in a training set of utterances paired with logical
forms that is used to train the semantic parser. 
We now briefly elaborate on these two steps.

\paragraph{Grammar} The grammar in \overnight{} generates logical forms paired
with canonical utterances, which are understandable to crowd workers (e.g., \nl{number of state that borders california}). 
The grammar has two parts: the \textit{domain-general} part contains
domain-independent rules that cover logical operators (e.g.,
comparatives, superlatives, negation etc.). In addition, a \textit{seed lexicon}
specifies a canonical phrase (\nl{publication year}) for each knowledge-base (KB) constant (\zl{PublicationYear}), in each particular domain.

While there are many possible ways to sample data from the grammar, in
\overnight{} logical forms and canonical utterances are exhaustively generated 
up to a certain maximal depth, hereafter denoted \md{}. This follows
the assumption that the semantic parser, trained from this data, should
generalize to logical forms that correspond to deeper trees (we re-examine this assumption in Section~\ref{sec:mismatch_analysis}).
In addition, a typing system is used during generation, thus, semantically vacuous
logical forms are not generated (e.g., \zl{PublicationYear.Parsing}), which
substantially reduces the number of generated examples.

\paragraph{Crowdsourcing}
After the above generation procedure terminates, crowd workers paraphrase each canonical utterance into a natural language utterance (e.g., the canonical utterance \nl{paper that has the largest publication year} can be paraphrased to \nl{what is the most recent paper?}). Finally, the framework yields 
a training dataset
$\don{}=\{(x_i,
z_i)\}_{i=1}^{N_{\text{on}}}$ consisting of pairs of natural language utterances and
logical forms, which can be used to train a semantic parser.

\section{Mismatch Analysis}
\label{sec:mismatch_analysis}

In  supervised semantic parsing, we assume access to training
data of the form $\dnat{}=\{(x_i,z_i)\}_{i=1}^{N_{\text{nat}}}$ sampled from
the true target distribution $p_{\text{nat}}(x,z)$. Conversely, in the \overnight{}
framework, we train from $\don$, which is sampled from a different distribution
$p_{\text{on}}(x,z)$. Training on data sampled from $p_{\text{on}}(x,z)$ and testing on data sampled
from $p_{\text{nat}}(x,z)$ leads to a substantial decrease in performance (compared to
training on data sampled from $p_{\text{nat}}(x,z)$). In this section, we will analyze and
quantify the causes for this degradation in performance.

By writing $p_{\text{nat}}(x,z) = p_{\text{nat}}(z)p_{\text{nat}}(x \mid z)$ and $p_{\text{on}}(x,z) = p_{\text{on}}(z)p_{\text{on}}(x \mid z)$ (mirroring the data generation procedure of \overnight{}), we can
decouple the distribution mismatch into two terms: The first is the
\emph{logical form mismatch}, i.e., the difference between the natural
distribution of logical forms $p_{\text{nat}}(z)$ and the distribution $p_{\text{on}}(z)$ of logical forms
induced by \overnight{}. The second is the \emph{language mismatch}, i.e., the
difference between the conditional distribution of natural language $p_{\text{nat}}(x \mid z)$
and the conditional distribution $p_{\text{on}}(x \mid z)$ of natural language when performing
crowdsourcing with the \overnight{} procedure. We will now investigate these two
types of mismatch and their interaction with neural semantic parsers in the
context of two popular semantic parsing datasets, \geo{} \cite{zelle96geoquery} and
\scholar{}
\cite{iyer2017neural}, which focus on the domains of geography and publications, respectively.
For this analysis, we replicated the \overnight{} procedure and generated a
dataset $\don$ for these two domains (details in
Section~\ref{sec:experiments}).

\subsection{Logical Form Mismatch}

\begin{table*}[t]
\centering
\resizebox{0.7\textwidth}{!}{
\begin{tabular}{llllll}
\\ \hline \hline
\multirow{2}{*}{operator} & \multicolumn{2}{c}{\geo{}} & \multicolumn{2}{c}{\scholar{}} & \multirow{2}{*}{example} \\
                          & $\dnat{}$      & $\don{}$       & $\dnat{}$         & $\don{}$        &                          \\ \hline
argmax      & .27         & .14     & .18           & .08       & \nl{the longest river}                         \\
argmin      & .10         & .14     & .02           & .08       & \nl{earliest paper by...}                      \\
larger      & .01         & .06     & .02           & .10       & \nl{high point higher than that of...}           \\
smaller    & .00         & .06     & .02           & .10       & \nl{paper with less than 10 citations}        \\
conj$_1$                   & .81         & .20     & .50           & .22       & \nl{parsing papers}                   \\
conj$_2$                   & .19         & .80     & .42           & .78       & \nl{2019 parsing papers}              \\
conj$_3$                   & .00         & .00     & .06           & .00       & \nl{2019 EMNLP parsing papers}        \\
conj$_4$                   & .00         & .00     & .01           & .00       & \nl{2019 EMNLP parsing papers by...} \\
negation                  & .01         & .55     & .01           & .57       & \nl{state with no rivers}              \\
aggregation               & .01         & .04     & .01           & .03       & \nl{total population of all states}      \\
count                     & .10         & .02     & .16           & .02       & \nl{number of papers published by...}      \\ \hline \hline
\end{tabular}} 
\caption{Logical operators frequency within the total number of examples. conj$_n$ refers to a logical from that contains a conjunction of $n$ clauses.}
\label{tab:log_op}
\end{table*}

The first source of mismatch is that the natural distribution of
logical forms $p_{\text{nat}}(z)$ can be quite different from the distribution $p_{\text{on}}(z)$ induced
by the ad-hoc procedure for generating logical forms in \overnight. 

\paragraph{Logical Operators Frequency}
A simple way to quantify the logical form mismatch, is to look whether the
proportion of different logical operators, such as superlatives (e.g.,
\zl{argmax}) and comparatives (e.g., \zl{>}), substantially differs between
$\don{}$ and $\dnat{}$. Table \ref{tab:log_op}
examines this, 
where hereafter we use the original training and development sets of \geo{} and \scholar{} as $\dnat{}$. The table shows that the frequency of
logical operators is different between the two training sets $\dnat{}$ and $\don{}$. For example,
in $\dnat{}$ of \geo{},
only $1\%$ of the
examples involve negation, compared to $55\%$ of the examples in $\don{}$. Such differences may present generalization difficulties for a model trained from $\don{}$, as 
the real distribution we care about is $\dnat{}$.   

\paragraph{Coverage}
\newcite{wang2015overnight} proposed to exhaustively generate all logical forms
corresponding to a derivation tree of maximal depth $D$, assuming that the model
will generalize to logical forms that require deeper trees. However, recent work
from \newcite{finegan-dollak-etal-2018-improving} showed this might not be the case. 
In their work, \newcite{finegan-dollak-etal-2018-improving} consider logical form
\emph{templates}, i.e., logical forms where the KB constants are
abstracted to their semantic type, and split the data such that templates in the test set
are disjoint from templates in the training set. They show that neural semantic
parsers struggle to generalize in this setup. Thus, \emph{covering} the logical form
templates that appear in $\dnat$ is important in the \overnight{} procedure.

Table \ref{tab:coverage} shows the number of examples generated by \overnight{}
for different values of \md{}, along with the 
proportion of examples in $\dnat$ whose logical form template is covered by $\don$.  
Because \overnight{} requires paraphrasing each generated example, a reasonable
value for \md{} is $5$. We observe that in this case only $63.73\%$ of the examples in \scholar{} are covered, as well as $76.52\%$ of the examples in \geo{}.

\begin{table}[t]
\centering
\resizebox{0.9\columnwidth}{!}{
\begin{tabular}{lccc} \hline\hline
Dataset                  & \md{} & $|\don{}|$ & $\dnat{}$ coverage \\ \hline
\multirow{4}{*}{\scholar{}} & 7          & 655,516       & 95.16          \\
                         & 6          & 91,739        & 90.85          \\
                         & 5          & 2,283             & 63.73               \\
                         & 4          & 91              & 17.44                \\ \hline
\multirow{4}{*}{\geo{}}     & 7          & 2,086,830              & 94.13                \\
                         & 6          & 249,617              & 89.43                \\
                         & 5          & 4,480             & 76.52               \\ 
                         & 4          & 166              & 51.68              \\ \hline\hline
\end{tabular}}
\caption{Logical form coverage of $\dnat{}$ by $\don{}$, after converting logical forms to their templates, for different values of \md{}.
}
\label{tab:coverage}
\end{table}

To verify whether coverage is indeed important in our setup, we performed the following experiment.
We trained a semantic parser (detailed in Section \ref{sec:experiments}) on $\don$ for \geo{} and \scholar{} (with $\md=5$), and evaluated
its performance on the training set of $\dnat$ (which serves as a development set
in this context).
Then, we calculated the accuracy of the model with respect to examples for which
their template appears in $\don{}$ ($acc_{\text{cov}}$), and for those their
template does not appear ($acc_{\text{disj}}$). Table \ref{tab:generalize} shows that
$acc_{\text{disj}}$ is substantially lower than $acc_{\text{cov}}$ for both datasets. 
This shows
the importance of generating logical form templates that are likely to
appear in the target distribution.
Thus, 
our finding reinforces the conclusions in \cite{finegan-dollak-etal-2018-improving}, and shows that while neural semantic parsers obtain good performance on existing benchmarks, their generalization to new compositional structures is limited.

\begin{table}[t]
\centering
\resizebox{0.6\columnwidth}{!}{
\begin{tabular}{lll} \hline \hline
Dataset & $acc_{\text{cov}}$ & $acc_{\text{disj}}$ \\ \hline
\geo{}     & 78.73  & 10.71   \\
\scholar{} & 65.58        & 5.24       \\ \hline \hline 
\end{tabular}}
\caption{Denotation accuracy for the set of examples in $\dnat{}$ covered by $\don{}$ ($acc_{\text{cov}}$), and for the set of uncovered examples ($acc_{\text{disj}}$).}
\label{tab:generalize}
\end{table}

\paragraph{Unlikely Logical Forms}
In \overnight{}, logical forms that are unlikely to appear in a natural
interaction could be generated.
For instance, the canonical utterance
\nl{total publication year of paper whose keyphrase is semantic parsing}, which
refers to the sum of publication years of all semantic parsing papers. Although
such logical forms are valid with respect to type matching, users are unlikely
to produce them, and they are hard to prune automatically. To estimate the
logical form mismatch caused by unlikely logical forms, we manually inspected
$100$ random examples from $\don{}$ with $\md=5$ for \geo{}, and found that 31\%
of the examples are unlikely. While these examples are not necessarily harmful
for a model, they are difficult to paraphrase, and may
introduce noisy paraphrases that hurt performance.

\subsection{Language mismatch}
\label{sec:lang_mis}

The second mismatch in \overnight{}, demonstrated in Figure \ref{fig:paraphrases}, is between the language used when naturally interacting with a conversational interface, and the language crowd workers use when elicited to generate paraphrases conditioned on a canonical utterance.

\begin{figure}[t]
  {\footnotesize
	\setlength{\fboxsep}{8pt}
	\fbox{
	\parbox{0.9\columnwidth}{
		
	\textbf{Example 1}
		
    $c$: \nl{river that traverses california and that has the largest length}
    
    \vspace{0.1cm}
    
    $x_\text{on}$: \nl{What river that traverses California has the largest length?}
    \vspace{0.1cm}
    
    $x_\text{nat}$: \nl{What is the longest river in California?}
    
	\vspace{0.2cm}
		
	\textbf{Example 2}
		
    $c$: \nl{paper whose keyphrase is deep learning and that has the largest publication year}
    
    \vspace{0.1cm}
    
    $x_\text{on}$: \nl{Name the paper with a deep learning keyphrase that has the most recent publication year.}
    
    \vspace{0.1cm}
    
    $x_\text{nat}$: \nl{Most recent deep learning papers.}
		
		}}
  }
  \caption{Examples for canonical utterances ($c$) generated by the grammar, their paraphrase by crowd workers ($x_\text{on}$), and their natural utterance in $\dnat$ ($x_\text{nat}$). While the paraphrases are correct, they are biased towards the language style in $c$.}
	\label{fig:paraphrases}
\end{figure}

To directly measure the language mismatch, we performed the following experiment. We generated a dataset $\dlang$ by
taking the examples from $\dnat$ and paraphrasing their logical forms using the \overnight{} procedure. This ensures that $p_\text{lang}(z)=p_\text{nat}(z)$, and thus the only difference between $\dnat$ and $\dlang$ is due to \emph{language mismatch}. Then we measured the difference in performance when training on these two datasets.

In more detail, for every example in $\dnat{}$, we extracted the corresponding canonical utterance template (since the examples are covered by the \overnight{} grammar), denoted by $\cnat{}$. This list of canonical utterances can be viewed as an oracle output of the \overnight{} grammar generation procedure, since they do not exhibit any logical form mismatch.
Next, 12 NLP graduate students paraphrased the examples in $\cnat{}$. 
The students were presented with guidelines and examples for creating paraphrases, similar to the original guidelines in \newcite{wang2015overnight}. Moreover, we explicitly asked to paraphrase the canonical utterances such that the output is significantly different from the input (while preserving the meaning). 
After the paraphrase process was completed, we manually fixed typos in all generated paraphrases. Thus, our paraphrasing procedure yields high quality paraphrases and is an upper bound to what can be achieved by crowd workers.

We trained a neural semantic parser on both $\dnat$ (named \textsc{Supervised}) and $\dlang$ (named \textsc{Overnight-oracle-lf}) using the original train/development split. Results in Table \ref{tab:language_mismatch} show that for each domain, a decrease of approximately 9 points in accuracy occurs only due to the language mismatch, even with high-quality workers. This gap is likely to grow when workers are not experienced in paraphrasing, or are unfamiliar with the domain they generate paraphrases for.

We additionally observe that when we use GloVe embeddings \cite{pennington2014glove} instead of contextual ELMo embeddings \cite{peters2018elmo}, the gap is even higher. This shows that better representations reduce the language gap, though it is still substantial.

\begin{table}[t]
\centering
\resizebox{0.98\columnwidth}{!}{
\begin{tabular}{lcc} \hline\hline
Model               & \geo{}   & \scholar{} \\ \hline
\textsc{Supervised+ELMo}          & 86.3 & 83.4   \\
\textsc{Overnight-oracle-lf+ELMo} & 77.7 & 73.5         \\
\textsc{Supervised+GloVe}          & 84.9 & 78.2   \\
\textsc{Overnight-oracle-lf+GloVe} & 71.6 & 67.2        \\ \hline \hline
\end{tabular}}
\caption{Denotation accuracy on the test set, comparing a semantic parser trained on $\dnat{}$ (\textsc{Supervised}) and parser trained on $\dlang{}$ (\textsc{Overnight-oracle-lf}).}
\label{tab:language_mismatch}
\end{table}

\section{Grammar-driven Annotation}
\label{grammar_driven_annotation}

To overcome the logical form and language mismatches discussed, we propose in this section a new data generation procedure that does not suffer from these shortcomings. Our procedure, named \detect{}, relies on two assumptions. First, unlike \overnight{}, we assume access to unlabeled utterances $\xul=\{x_i\}_{i=1}^{N_{\text{ul}}}$. These can typically be found in query logs, or generated by users experimenting with a prototype. Second, we assume a scoring function $s_0(x, c)$, which provides a reasonable initial similarity score between a natural language utterance $x$ and a canonical utterance $c$.

The goal of our procedure is to iteratively label $\xul$ with crowd workers, aided by the \overnight{} grammar.
If we manage to label most of the dataset $\xul$, which is sampled from the true target distribution,
we will end up with a labeled dataset $\dga{}$ that has very little distribution mismatch.

Figure~\ref{fig:detect} illustrates the procedure. First, we generate all canonical utterances $\con$ up to a depth $D$ from the \overnight{} grammar. Because we do not paraphrase all the canonical utterances, we can generate to a higher depth $D$ compared to \overnight{} and cover more of the examples in $\xul$ (see Table~\ref{tab:coverage}). Then, we iteratively label the utterances in $\xul$. At each iteration $t$, we use a paraphrase detection model $s_t(x, c)$ to present promising canonical utterances to crowd workers for each unlabeled utterance $x$, who label the dataset. Importantly, crowd workers in our setup do not \emph{generate paraphrases}, they only \emph{detect} them. We now describe \detect{} in more detail.

\paragraph{Iterative Annotation}
At each iteration $t$, we rely on a trained scoring function $s_t(x,c)$ that estimates the similarity between an unlabeled utterance $x \in \xul{}$ and a generated canonical utterance $c \in \con{}$. We follow the procedure described in Algorithm~\ref{alg:detect}. For an utterance $x \in \xul{}$, we calculate the top-$K$ ($=5$) most similar canonical utterances in $\con{}$, denoted by $\sC^{K}_x$. We then present $x$ along with its candidate paraphrases $\sC^{K}_x$ to a worker, and ask her to choose the correct candidate paraphrase. If a paraphrase does not appear in the top-$K$ candidates, the worker selects no candidates. 

\begin{algorithm}[t]
  {\footnotesize
\caption{\detect{}}\label{alg:detect}
\begin{algorithmic}[1]
  \Require unlabeled utterances $\xul{}$, grammar $G$
\Ensure $\dga{}$ - training data for a semantic parser
  \State generate $\con{}$ from $G$
  \State $\dga{} \gets \emptyset$, $s_0(\cdot) \gets -WMD(\cdot)$, $t \gets 0$ 
  \State $\emph{converge} \gets False$, $\sS_{\text{pos}} \gets \emptyset$, $\sS_{\text{neg}} \gets \emptyset$
\While  {\textbf{not } $\emph{converge}$}
\For{$x$ \textbf{in} $\xul{}$}
  \State calculate $\sC^{K}_{x}$ using $s_t(\cdot)$ \Comment{top $K$ candidates}
  \State crowd workers annotate $c_{x}$ given $x$ and
  $\sC^{K}_{x}$
  \If{$c_{x} \neq N/A$}
    \State $\dga{} \gets \dga{} \cup (x,c_x)$
    \State $\xul{} \gets \xul{} \setminus x$
    \State $\sS_{\text{pos}} \gets \sS_{\text{pos}} \cup (x,c_x)$ \Comment{positive examples}
    \For{$c$ \textbf{in} $\sC^{M}_x \setminus \{c_{x}\}$}
    \State $\sS_{\text{neg}} \gets \sS_{\text{neg}} \cup (x,c)$ \Comment{negative examples}
      \EndFor
  \EndIf
\EndFor
\State $\emph{converge} \gets$ check for convergence, $t \gets t+1$
\State train $s_{t}(\cdot)$ over $\sS_{\text{pos}}$ and $\sS_{\text{neg}}$
\EndWhile
\State \textbf{return} $\dga{}$
\end{algorithmic}
  }
\end{algorithm}

These annotations are then used to train $s_{t+1}(x, c)$, which will be used in the next iteration.
For each $x\in \xul{}$ for which a worker detected a paraphrase $c_{x}$, we label $(x, c_x)$ as a positive example. We use the top-$M$ ($=100$) other most similar canonical utterances $\sC^{M}_x \setminus \{c_{x}\}$ (according to $s_t(\cdot)$) as negative examples. We train $s_{t+1}$ from all the examples generated in iterations $0 \cdots t$.
Thus, in every iteration more and more examples are labeled, and a better scoring function is trained. We stop when we meet convergence, i.e., when no new unlabeled utterances are labeled. We then output the dataset $\dga{}$ that contains every utterance $x$ from $\xul{}$ that is now labeled, paired with the logical form that corresponds to the detected canonical utterance paraphrase $c_x$. 

We note that an alternative modeling choice was to use the semantic parser itself as the scorer for candidate canonical utterances. However, decoupling the parser and the scorer is beneficial, as the discriminative scoring function $s_t(\cdot)$ benefits from negative examples (incorrect paraphrases), unlike the generative semantic parser.

The success of our procedure depends on a good initial scoring function $s_0(x, c)$, to be used in the first iteration, that we next describe.

\paragraph{Initial Scoring Function}
We implement $s_0(x,c)$ in an unsupervised manner, as no labeled examples are available in the first iteration.
Formally, we take $s_0(x,c)=-WMD(x,c)$, where $WMD(x,c)$ is the Word Mover's Distance (WMD) \cite{kusner2015word} between $x$ and $c$, which is the minimum amount of distance that
the embedded words of one utterance need to travel to reach those of the other utterance.
We found WMD to perform better than cosine similarity over averaged pre-trained embeddings, as WMD performs word-level alignment, and shared words (such as entities) encourage small distance.

\paragraph{Implementation details}
We take the unlabeled utterance set to be all utterances in $\dnat{}$, when ignoring logical forms: $\xul{}=\{x \mid (x,\cdot) \in \dnat{}\}$. We generate all canonical utterances up to depth $\md=6$, resulting in roughly 350K canonical utterances in \scholar{} and \geo{}, and coverage of 90\% of the examples in $\xul$ (Table~\ref{tab:coverage}).

Our binary classifier $s_t(\cdot)$ is trained from paraphrases detected by workers. We utilize the \textsc{ESIM} neural architecture originally used for natural language inference \cite{chen-etal-2017-enhanced}. We also employ ELMo contextualized embeddings \cite{peters2018elmo}, and minimize the binary cross-entropy loss for each example.
\section{Experiments}
\label{sec:experiments}

\subsection{Experimental Setup}

\paragraph{Datasets}
We experiment with two popular semantic parsing datasets: \geo{} \cite{zettlemoyer05ccg} and \scholar{} \cite{iyer2017neural}, that contain questions about US geography and academic publications, respectively. 

Because we utilize the original grammar from \newcite{wang2015overnight} that generates logical forms in lambda-DCS \cite{liang2013lambdadcs}, we first manually annotated \geo{} and \scholar{} with lambda-DCS logical forms that are translations of the original logical forms (Prolog for \geo{} and SQL for \scholar{}). We only convert examples that are covered by the \overnight{} grammar, which results in annotating $99.3\%$ of the examples in \geo{} (874 in total), and $96.7\%$ of the examples in \scholar{} (790 in total). 

\paragraph{Grammar Generation}
To generate the training data $\don{}$ for our \overnight{} baseline and \detect{}, we first exhaustively generated logical form and canonical utterance pairs from the grammar up to depth $\md{}=5$ for \overnight{} and $\md{}=6$ for \detect{}, using type matching rules such that vacuous logical forms are not generated. Then, we further pruned unlikely logical forms that can be automatically detected (e.g., contradictions such as \nl{state that borders california and that not borders california}). We additionally pruned equivalent examples: if we generated a logical form with the structure \zl{A $\sqcap$ B}, we pruned \zl{B $\sqcap$ A}.

\paragraph{Crowd Sourcing}
We gathered annotations from crowd workers by running a qualification task where we manually verified workers (annotators with at least 85\% success rate were qualified). Then, qualified workers performed the tasks for the remaining examples, and for those where unqualified workers failed. 

For \overnight{}, we gathered a single paraphrase per canonical utterance in $\don{}$ with a cost of 0.06\$. For \detect{}, we detected a single paraphrase for each unlabeled utterance in $\xul{}$ that was not annotated in previous iterations, with a cost of 0.05\$. 

In total, we gathered 7,140 paraphrases for \overnight{} with a total cost of 515\$, and 2,594 detections for \detect{} with a total cost of 155\$. Thus, \detect{} had lower cost per task, and given the results below, benefits more from fewer tasks.

\paragraph{Neural Semantic Parser}
We use a standard semantic parser provided by AllenNLP \cite{Gardner2017AllenNLP}, based on a sequence-to-sequence model \cite{sutskever2014sequence}. The encoder is a BiLSTM that receives ELMo pre-trained embeddings \cite{peters2018elmo} as input. The attention-based decoder \cite{bahdanau2015neural} is an LSTM that also employs copying \cite{jia2016recombination}. 

\paragraph{Training Scheme}
For both the semantic parser and the paraphrase detection model $s_t(\cdot)$ we take $10\%$ of the training data for validation (or the official development set, if it exists). When training both models, we abstract examples to their templates, as in \newcite{dong2016logical} and \newcite{finegan-dollak-etal-2018-improving} inter alia. 
For our semantic parser, we tune the learning rate and dropout over the development set, and for $s_t(\cdot)$ we use the same hyper-parameter values as in \newcite{chen-etal-2017-enhanced}. We use early-stopping, and choose the model with the highest denotation accuracy and $F_1$ measure for the semantic parser and $s_t(\cdot)$, respectively.

\subsection{Results}

\begin{table}[t]
\centering
\resizebox{0.92\columnwidth}{!}{
\begin{tabular}{lcc} \hline\hline
Model               & \geo{}   & \scholar{} \\ \hline
\textsc{Supervised}          & 86.3 & 83.4   \\
\detect{}               & 72.0 & 69.2       \\
\overnight{}           & 61.9 & 40.8   \\
\hline
\overnight{}\textsc{-oracle-lf} & 77.7 & 73.5         \\
\detect{}\textsc{-orcale}         & 79.1 & 72.5   \\ \hline\hline
\end{tabular}}
\caption{Denotation accuracy on the test set.}
\label{tab:results}
\end{table}

\paragraph{Main Results}
Table \ref{tab:results} shows the denotation accuracy for all experiments, when training from: $\dnat{}$ (\textsc{Supervised}); $\don{}$ (\overnight{}); $\dlang$, as described in Section \ref{sec:lang_mis} (\overnight{}\textsc{-oracle-lf}); $\dga{}$, described in Section \ref{grammar_driven_annotation} (\detect{}); and $\dga{}$ where we simulate a perfect worker that always detects the gold paraphrase if it appears in the top-$K$ candidates (\detect{}\textsc{-orcale}).

\begin{table*}[t]
\centering
\resizebox{1.0\textwidth}{!}{
\begin{tabular}{lllll}
\hline \hline
Case & \% & Natural language utterance                            & Gold canonical utterance                        & Detected canonical utterance                   \\ \hline
\emph{us}   & .48   & \nl{what is the highest point in california?} & \nl{place that is high point of california} & \nl{high point of california}             \\
\emph{os}   & .13   & \nl{how many citizens in california?}            & \nl{population of california}              & \nl{total population of california}       \\
\emph{w}    & .08   & \nl{which state has the greatest density?}       & \nl{state that has the largest density}    & \nl{total density of state}               \\
\emph{pw}   & .31   & \nl{what is the lowest elevation in california?} & \nl{place that is low point of california} & \nl{elevation of low point of california} \\ \hline \hline
\end{tabular}}
\caption{Examples of false positive detections by crowd workers for different cases: \emph{under specification} (\emph{us}), \emph{over specification} (\emph{os}), \emph{wrong} (\emph{w}) and \emph{partially wrong} (\emph{pw}).
}
\label{tab:false_pos}
\end{table*}

Results show that \overnight{} achieves substantially lower accuracy compared to training with examples from the target distribution (\textsc{Supervised}), inline with the analyses presented in Section \ref{sec:mismatch_analysis}. For instance, \scholar{} accuracy more than halves (40.8 for \overnight{} in comparison to 83.4 for \textsc{Supervised}).
Conversely, our suggested method \detect{} that directly annotates unlabeled utterances achieves much higher accuracy than \overnight{}. Utilizing crowd workers for detection leads to 70.6 accuracy on average, and to 75.8 when simulating perfect crowd workers (\detect{}\textsc{-oracle}).

All models performed well on the development set that was sampled from the same distribution as the training set ($>80\%$ accuracy), and thus differences in performance are due to generalization to the true distribution.
Moreover, the accuracy of the paraphrase detection model $s_t(\cdot)$ was very high ($>95\%$ $F_1$ measure) on the development set, showing that detection is easier to model compared to generation.

We note in passing that we also implemented a baseline that uses unlabeled examples in conjunction with \overnight{} through self-training \cite{konstas2017neural}. We trained a model with \overnight{} and iteratively labeled unlabeled utterances for which model confidence was high. However, we were unable to obtain good performance with this method.

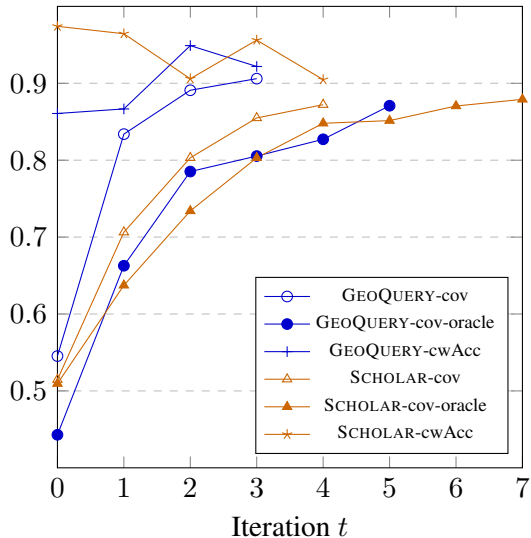
\begin{figure}[t]
\begin{tikzpicture}
\begin{axis}[
	width = 1.0\columnwidth,
	height = 1.0\columnwidth,
    xlabel={Iteration $t$},
    xmin=0, xmax=7,
    ymin=0.4, ymax=1,
    xtick={0, 1, 2, 3, 4, 5, 6, 7},
    ytick={0.5, 0.6, 0.7, 0.8, 0.9},
    legend pos=south east,
    legend style={font=\scriptsize},
    %cycle list name=exotic,
    ymajorgrids=true,
    grid style=dashed,
]

\addplot
	[
    color=blue!80!black,
    mark=o,
    ]
    coordinates {
    (0,0.545302013)(1,0.833892617)(2,0.890939597)(3,0.906040268)
    };
    \addlegendentry{\geo{}-cov} 
\addplot
	[
    color=blue!80!black,
    mark=*,
    ]
    coordinates {
    (0,0.44295302)(1,0.662751678)(2,0.785234899)(3,0.805369128)(4,0.827181208)(5,0.870805369)
    };
    \addlegendentry{\geo{}-cov-oracle}
\addplot
	[
    color=blue!80!black,
    mark=+,
    ]
    coordinates {
    (0,0.860738255033557)(1,0.866666666666666)(2,0.948979591836734)(3,0.921875)
    };
    \addlegendentry{\geo{}-cwAcc}
\addplot
	[
    color=orange!80!black,
    mark=triangle,
    ]
    coordinates {
    (0,0.514680484)(1,0.706390328)(2,0.803108808)(3,0.85492228)
    (4,0.872193437)
    };
    \addlegendentry{\scholar{}-cov}
\addplot
	[
    color=orange!80!black,
    mark=triangle*,
    ]
    coordinates {
    (0,0.509499136)(1,0.637305699)(2,0.73402418)(3,0.803108808)
    (4,0.848013817)(5,0.851468048)(6,0.870466321)(7,0.8791019)
    };
    \addlegendentry{\scholar{}-cov-oracle}
\addplot
	[
    color=orange!80!black,
    mark=star,
    ]
    coordinates {
    (0,0.974093264)(1,0.964412811)(2,0.905882353)(3,0.956140351)
    (4,0.904761905)
    };
    \addlegendentry{\scholar{}-cwAcc}
\end{axis}
\end{tikzpicture}
\caption{Analysis of \detect{} across iterations.}
\label{fig:cov_curve}
\end{figure}

\paragraph{\detect{} Analysis}
\detect{} iteratively annotates unlabeled utterances utilizing crowd workers. Figure \ref{fig:cov_curve} reports several metrics for \detect{} for each iteration: \emph{cov} details the fraction of annotated utterances in $\xul{}$, where \emph{cov-oracle} corresponds to the annotation coverage by \detect{}\textsc{-oracle}. In addition, cwAcc is the crowd workers' detection accuracy per iteration, with respect to the gold canonical utterances. The figure shows that in both datasets, \detect{} converges in a few iterations, and that workers' detection accuracy is high across all iterations ($>85\%$). 

An interesting phenomenon is that crowd workers (\detect{}) cover the unlabeled utterances faster than oracle workers (\detect{}\textsc{-oracle}). 
To analyze this, we inspect false positive, i.e., cases where the gold canonical utterance does \emph{not} appear in the top-$K$ candidates of $s_t(\cdot)$, but the crowd worker detects some candidate as the paraphrase. 
Table \ref{tab:false_pos} presents examples for these cases and their fraction within all false positives for iteration $t=0$, where workers cover unlabeled utterances faster than the oracle.
We find that in 61\% of the cases, 
the choice of the workers was equivalent to the gold candidate.
This is due to \emph{under specification}, when the gold paraphrase is more specific than the detected one, or \emph{over specification}, which is the opposite case. The other 39\% are indeed errors, which we split into \emph{wrong} detections, and \emph{partially wrong} detections, where the detected paraphrase is different than the gold one, but is reasonable choice the phrasing of the question. E.g., for \nl{what is the lowest elevation in California} it is unclear whether the answer should be a location or the elevation. This shows that many false positives are in fact correct.

\comment{
Although this gap might be due to cases where workers annotate an unlabeled utterance with some canonical utterance candidate although it is the wrong one, we find multiple examples where the annotation is still correct. We categorize two common cases: \emph{Under specification} occurs when the gold logical form is more specified than the detected logical form, while they are both correct, e.g.,\nl{city that is capital of california} in comparison to \nl{capital of california}. \emph{Over specification} is the opposite case, e.g., when the gold logical form is \nl{author that writes paper whose keyphrase is semantic parsing}, and the detected logical form is \nl{author that writes paper and ....}
}

\paragraph{Limitations}
\detect{} relies on the unsupervised function $s_0(\cdot)$ to bootstrap the annotation procedure. In both datasets, $s_0(\cdot)$ managed to rank gold paraphrases within its top-$5$ candidates for roughly half of the unlabeled utterances in $\xul{}$, but this is not guaranteed.

During each iteration in \detect{}, the function $s_t(\cdot)$ is applied on all pairs of an unlabeled utterance and candidate canonical utterances, thus $s_t(\cdot)$ is applied $O(|\xul{}| \cdot |\don{}|)$ times. Empirically, we find that computation time is manageable when limiting the application of $s_t(\cdot)$ to candidates that share the same entities as the unlabeled utterance. However, this might not suffice for KBs with large schemas.
In such cases, an information retrieval module could retrieve a small number of candidates, similar to \newcite{yang2019end}.
\section{Related Work}

Several works used a human in the loop for training a semantic parser. \newcite{iyer2017neural} incorporated user feedback to detect wrong parses and sent them to expert annotation.
\newcite{lawrence-riezler-2018-improving} and \newcite{berant2018explaining} improved a supervised parser by showing its predictions to users and training from this feedback.
\newcite{gupta-etal-2018-semantic-parsing} built a hierarchical annotation scheme for annotating utterances with multiple intents.
\newcite{labutov2019learning} trained a semantic parser through interactive dialogues.
Comparing to these works, our proposed method requires no supervised data or expert annotators, and is suitable for rapid development of parsers in multiple domains.

In semi-supervised learning, \newcite{konstas2017neural} used self-training to improve an existing AMR parser. 
Others used a variational auto-encoder by modeling unlabeled utterances \cite{kocisk2016semantic} or logical forms \cite{yin-etal-2018-structvae} as latent variables.
However, empirical gains from the unlabeled data were relatively small compared to annotating more examples.

Finally, several papers extended the \overnight{} procedure. \newcite{ravichander-etal-2017-say} replaced phrases in the lexicon with images to elicit more natural language from workers. \cite{cheng2018building} 
generated more complex compositional structures by splitting the canonical utterances into multiple steps. 
Such work relies on workers to  \emph{generate} paraphrases, while we propose to simply \emph{detect} them.
\section{Conclusion}
We address the challenge of generating data for training semantic parsers from scratch in multiple domains.
We thoroughly analyze the \overnight{} procedure, and shed light on the factors that lead to poor generalization, namely logical form mismatch and language mismatch. We then propose \detect{}, a method that directly annotates unlabeled utterances with their logical form, by letting crowd workers detect automatically-generated canonical utterances. We demonstrate our method's success on two popular datasets, and find it substantially improves generalization to real data, compared to \overnight{}.

\section*{Acknowledgments}
We thank the anonymous reviewers for their constructive feedback. This work was completed in partial fulfillment for the PhD degree of the first author, which was also supported by a Google PhD fellowship. This research was partially supported by The Israel Science Foundation grant 942/16, The Blavatnik Computer Science Research Fund and The Yandex Initiative for Machine Learning.

\bibliography{all}
\bibliographystyle{acl_natbib}

\end{document}